\documentclass[11pt,letterpaper]{article}
\usepackage{times}
\usepackage{latexsym}
\usepackage{amsmath}

\title{Skip-gram Language Modeling Using Sparse Non-negative Matrix Probability Estimation}

\author{Noam Shazeer \and Joris Pelemans \and Ciprian Chelba\\
  Google, Inc., 1600 Amphitheatre Parkway\\
  Mountain View, CA 94043, USA\\
  {\tt \{noam,jpeleman,ciprianchelba\}@google.com}}

\date{}

\begin{document}
\maketitle
\begin{abstract}
We present a novel family of language model (LM) estimation techniques named 
Sparse Non-negative Matrix (SNM) estimation. 

A first set of experiments empirically evaluating it on the One Billion Word 
Benchmark~\cite{Chelba:2013} shows that SNM $n$-gram LMs perform almost as well
as the well-established Kneser-Ney (KN) models. 
When using skip-gram features the models are able to match the state-of-the-art
recurrent neural network (RNN) LMs; combining the two modeling techniques yields
the best known result on the benchmark.

The computational advantages of SNM over both maximum entropy and RNN LM estimation 
are probably its main strength, promising an approach that has the same flexibility
in combining arbitrary features effectively and yet should scale to very large
amounts of data as gracefully as $n$-gram LMs do.
\end{abstract}

\section{Introduction}
\label{sec:intro}

A statistical language model estimates probability values $P(W)$ for
strings of words $W$ in a vocabulary ${\mathcal V}$ whose size is in the
tens, hundreds of thousands and sometimes even millions.
Typically the string $W$ is broken into sentences, or other segments such 
as utterances in automatic speech recognition, which are often assumed to be
conditionally independent; we will assume that $W$ is such a segment, 
or sentence.

Estimating full sentence language models is computationally hard if
one seeks a properly normalized probability 
model\footnote{We note that in some
  practical systems the constraint on using a properly normalized
  language model is side-stepped at a gain in modeling power and
  simplicity.} over strings of
words of finite length in ${\cal V}^*$. 
A simple and sufficient way to ensure proper normalization of the
model is to decompose the sentence probability according to the chain
rule and make sure that the end-of-sentence symbol \verb+</s>+ is
predicted with non-zero probability in any context. With
$W=w_1,w_2,\ldots,w_n$ we get:
\begin{eqnarray}
  \label{intro:bayes}
  P(W)=\prod_{i=1}^nP(w_i|w_1,w_2,\ldots,w_{i-1}) 
\end{eqnarray}

Since the parameter space of $P(w_k|w_1,w_2,\ldots,w_{k-1})$ is too
large, the language model is forced to put the \emph{context}
$W_{k-1}=w_1,w_2,\ldots,w_{k-1}$ into an \emph{equivalence class} determined
by a function $\Phi(W_{k-1})$. As a result,
\begin{equation}
\label{e1}P(W)\cong\prod_{k=1}^nP(w_k|\Phi (W_{k-1})) 
\end{equation}

Research in language modeling consists of finding appropriate
equivalence classifiers $\Phi$ and methods to estimate
$P(w_k|\Phi(W_{k-1}))$. The most successful paradigm in language modeling uses the
\emph{$(n-1)$-gram} equivalence classification, that is, defines%
$$
\Phi (W_{k-1})\doteq w_{k-n+1},w_{k-n+2},\ldots,w_{k-1}
$$
Once the form $\Phi (W_{k-1})$ is specified, only the problem of
estimating $P(w_k|\Phi (W_{k-1}))$ from training data remains.

\subsection*{Perplexity as a Measure of Language Model Quality}

A \emph{statistical language model} can be evaluated by how well it
predicts a string of symbols $W_t$---commonly referred to as
\emph{test data}---generated by the source to be modeled.

A commonly used quality measure for a given model $M$ is related to
the entropy of the underlying source and was introduced under
the name of \emph{perplexity} (PPL):
\begin{eqnarray}
\label{basic_lm:ppl}
PPL(M) = exp(-\frac{1}{N} \sum_{k=1}^{N}\ln{[P_M(w_k|W_{k-1})]}) 
\end{eqnarray}

For an excellent discussion on the use of perplexity in 
statistical language modeling, as well as various estimates
for the entropy of English the reader is referred
to~\cite{jelinek97},~Section 8.4,~pages~141-142 and the additional reading
suggested in Section~8.5 of the same book.

Very likely, not all words in the test string $W_t$ are
part of the language model vocabulary. It is common practice to map
all words that are out-of-vocabulary to a distinguished \emph{unknown word}
symbol, and report the out-of-vocabulary (OOV) rate on test data---the
rate at which one encounters OOV words in the test string $W_t$---as
yet another language model performance metric besides
perplexity. Usually the unknown word is assumed to be part of the
language model vocabulary---\emph{open vocabulary} language models---and
its occurrences are counted in the language model perplexity
calculation, Eq.~(\ref{basic_lm:ppl}). A situation less common in
practice is that of \emph{closed vocabulary} language models where all
words in the test data will always be part of the vocabulary ${\cal  V}$.

\section{Skip-gram Language Modeling}
\label{sec:skip}

Recently, neural network (NN) smoothing~\cite{Bengio:2003},~\cite{Emami:2006},~\cite{Schwenk:2007}, and in particular recurrent neural networks~\cite{Mikolov:2012} (RNN) 
have shown excellent performance in language modeling~\cite{Chelba:2013}.
Their excellent performance is attributed to a combination of leveraging 
long-distance context, and training a vector representation for words.

Another simple way of leveraging long distance context is to use skip-grams.
In our approach, a skip-gram feature extracted from the context $W_{k-1}$ is characterized by the tuple $(r, s, a)$ where:
\begin{itemize}
\item $r$ denotes number of remote context words
\item $s$ denotes the number of skipped words
\item $a$ denotes the number of adjacent context words
\end{itemize}
relative to the target word $w_k$ being predicted.
For example, in the sentence, 
\verb+<S> The quick brown fox jumps over the lazy dog </S>+\\
a $(1, 2, 3)$ skip-gram feature for the target word \verb+dog+ is:\\
\verb+[brown skip-2 over the lazy]+

For performance reasons, it is recommended to limit $s$ and to limit
either $(r + a)$ or limit both $r$ and $s$; not setting any limits
will result in events containing a set of skip-gram features 
whose total representation size is quintic in the length of
the sentence.

We configure the skip-gram feature extractor to produce all features $\mathbf{f}$, defined by 
the equivalence class $\Phi(W_{k-1})$, that meet constraints on the minimum and maximum values for:
\begin{itemize}
\item the number of context words used $r + a$;
\item the number of remote words $r$;
\item the number of adjacent words $a$;
\item the skip length $s$.
\end{itemize}

We also allow the option of not including the exact value of $s$ in the 
feature representation; this may help with smoothing by sharing counts 
for various skip features. Tied skip-gram features will look like:\\
\verb+[curiousity skip-* the cat]+

In order to build a good probability estimate 
for the target word $w_k$ in a context $W_{k-1}$ we need a way of combining 
an arbitrary number of skip-gram features $\mathbf{f}_{k-1}$, which do not fall into a simple 
hierarchy like regular $n$-gram features. The following section describes 
a simple, yet novel approach for combining such predictors in a way that 
is computationally easy, scales up gracefully to large amounts of data 
and as it turns out is also very effective from a modeling 
point of view.

\section{Sparse Non-negative Matrix Modeling}
\label{sec:snm}
\subsection{Model definition}
In the Sparse Non-negative Matrix (SNM) paradigm, we represent the training data 
as a sequence of events $E = {e_1,e_2,...}$ where each event $e \in E$ consists 
of a sparse non-negative feature vector $\mathbf{f}$ and a sparse non-negative target word 
vector $\mathbf{t}$. Both vectors are binary-valued, indicating the presence or absence of a 
feature or target words, respectively. Hence, the training data consists of $|E||Pos(\mathbf{f})|$ 
positive and $|E||Pos(\mathbf{f})|(|\mathcal{V}|-1)$ negative training examples, where $Pos(\mathbf{f})$ 
denotes the number of positive elements in the vector $\mathbf{f}$.

A language model is represented by a non-negative matrix $\mathbf{M}$ that, when 
applied to a given feature vector $\mathbf{f}$, produces a dense prediction vector $\mathbf{y}$:
\begin{equation}
	\mathbf{y} = \mathbf{M} \mathbf{f} \approx \mathbf{t}
\end{equation}
Upon evaluation, we normalize $\mathbf{y}$ such that we end up with a conditional probability 
distribution $P_{\mathbf{M}}(\mathbf{t}|\mathbf{f})$ for a model $\mathbf{M}$. For each word $w \in \mathcal{V}$ that corresponds to 
index $j$ in $\mathbf{t}$, and its feature vector $\mathbf{f}$ that is defined by the equivalence class
$\Phi$ applied to the history $h(w)$ of that word in a text, the conditional probability $P_{\mathbf{M}}(w|\Phi(h(w)))$
then becomes:
%This means that for each $w_j$ in $\mathcal{V}$, the evaluation of Eq.~(\ref{e1}) boils down to:
\begin{equation}
\label{eq:prob}
%P(w_j|\Phi (h(w_j))) = \frac{\sum_{i \in Pos(\mathbf{f})}M_{ij}}{\sum_{i \in Pos(\mathbf{f})}\sum_{u=1}^{|\mathcal{V}|} M_{iu}}
	P_{\mathbf{M}}(w|\Phi (h(w))) = P_{\mathbf{M}}(t_j|\mathbf{f}) = \frac{y_j}{\sum_{u=1}^{|\mathcal{V}|} y_u} 
	= \frac{\sum_{i \in Pos(\mathbf{f})}M_{ij}}{\sum_{i \in Pos(\mathbf{f})}\sum_{u=1}^{|\mathcal{V}|} M_{iu}}
\end{equation}
%where $h(w_j)$ denotes the history of word $w_j$ in a given text and $\mathbf{f}$ is the feature vector defined by applying an 
%equivalence class function $\Phi$ to this history. 
For convenience, we will write $P(t_j|\mathbf{f})$ instead of $P_{\mathbf{M}}(t_j|\mathbf{f})$ in the rest of the 
paper.

As required by the denominator in Eq.~(\ref{eq:prob}), this computation involves summing over all of the present 
features for the entire vocabulary. However, if we precompute the row sums $\sum_{u=1}^{|\mathcal{V}|} M_{iu}$ and store them 
together with the model, the evaluation can be done very efficiently in only $|Pos(\mathbf{f})|$ time. Moreover, only the 
positive entries in $M_i$ need to be considered, making the range of the sum sparse.

\subsection{Adjustment function and metafeatures}
We let the entries of $\mathbf{M}$ be a slightly modified version of the relative frequencies:
\begin{equation}
	%M_{ij} = e^{A(f_i,t_j,C_{i*},C_{ij})} \frac{C_{ij}}{C_{i*}}
	M_{ij} = e^{A(i,j)} \frac{C_{ij}}{C_{i*}}
\end{equation}
where $\mathbf{C}$ is a feature-target count matrix, computed over the entire training corpus and $A(i,j)$ is 
a real-valued function, dubbed \emph{adjustment function}. For each feature-target pair $(f_i,t_j)$, the adjustment function 
extracts $k$ new features $\alpha_k$, called \emph{metafeatures}, which are hashed as keys 
to store corresponding weights $\theta(hash(\alpha_k))$ in a huge hash table. To limit memory usage, we use a flat hash table and allow 
collisions, although this has the potentially undesirable effect of tying together the weights of different metafeatures. 
Computing the adjustment function for any $(f_i,t_j)$ then amounts to summing the weights that correspond 
to its metafeatures:
\begin{equation}
	%A(f_i,t_j,C_{i*},C_{ij}) = \sum_k \theta_k \alpha_k(f_i,t_j,C_{i*},C_{ij})
	%A(i,j) = \sum_k \theta_k \delta(k, hash[\alpha_k(f_i,t_j,C_{i*},C_{ij})])
	%A(i,j) = \sum_k \theta(hash[\alpha_k(f_i,t_j,C_{i*},C_{ij})])
	A(i,j) = \sum_k \theta(hash[\alpha_k(i,j)])
\end{equation}
From the given input features, such as regular $n$-grams and skip $n$-grams, we construct our metafeatures as conjunctions of 
any or all of the following elementary metafeatures:
\begin{itemize}
	\item feature identity, e.g. \verb+[brown skip-2 over the lazy]+
	\item feature type, e.g. $(1, 2, 3)$ skip-grams
	\item feature count $C_{i*}$
	\item target identity, e.g. \verb+dog+
	\item feature-target count $C_{ij}$
\end{itemize}
where we reused the example from Section~\ref{sec:skip}. Note that the seemingly absent feature-target identity is represented by the conjunction of 
the feature identity and the target identity. Since the metafeatures may involve the feature count and feature-target count, 
in the rest of the paper we will write $\alpha_k(i,j,C_{i*},C_{ij})$. This will become important later when we discuss leave-one-out training.

Each elementary metafeature is joined with the others to form more complex metafeatures which in turn are joined with all the other 
elementary and complex metafeatures, ultimately ending up with all $2^5 - 1$ possible combinations of metafeatures. 

Before they are joined, count metafeatures are bucketed together according to their (floored) $\log_2$ value. As this effectively puts 
the lowest count values, of which there are many, into a different bucket, we optionally introduce a second (ceiled) bucket to assure smoother transitions. 
Both buckets are then weighted according to the $\log_2$ fraction lost by the corresponding rounding operation. Note that if we apply double 
bucketing to both the feature and feature-target count, the amount of metafeatures per input feature becomes $2^7 - 1$.
%\begin{enumerate}
%	\item $\lfloor \log_2(x) \rfloor, w_1 = \lfloor \log_2(x) \rfloor + 1 - \log_2(x)$ 
%	\item $\lfloor \log_2(x) \rfloor + 1, w_2 = \log_2(x) - \lfloor \log_2(x) \rfloor$
%\end{enumerate}

We will come back to these metafeatures in Section~\ref{sec:ablations} where we examine their individual effect on the model.

\subsection{Loss function}
Estimating a model $\mathbf{M}$ corresponds to finding optimal weights $\theta_k$ for all the metafeatures for all events in such a way 
that the average loss over all events between the target vector $\mathbf{t}$ and the prediction vector $\mathbf{y}$ is minimized, according to some loss function $L$. 
The most natural choice of loss function is one that is based on the multinomial distribution. That is, we consider $\mathbf{t}$ to be multinomially distributed 
with $|\mathcal{V}|$ possible outcomes. The loss function $L_{multi}$ then is:
\begin{equation}
	L_{multi}(\mathbf{y},\mathbf{t}) = - log(P_{multi}(\mathbf{t}|\mathbf{f})) = - log(\frac{y_j}{\sum_{u=1}^{|\mathcal{V}|}y_u}) = log(\sum_{u=1}^{|\mathcal{V}|}y_u) - log(y_j)
\end{equation}

%The likelihood of $t_j$ then equals:
%\begin{equation}
%	\mathcal{L}_{multi}(t_j) = \frac{\sum_{i \in \mathbf{f}} C_{ij} A(i,j)}{\sum_{u \in \mathcal{V}} \sum_{i \in \mathbf{f}} C_{iu} A(i,u)}
%\end{equation}
%and the corresponding loss function, over the entire training data $\mathcal{T}$, is:
%\begin{equation}
%	L_{multi}(\mathcal{T}) = \sum_{e \in E} log P(\mathbf{t}(e)|\mathbf{f}(e))
%\end{equation}
%where $E$ denotes the entire set of training events.

Another possibility is the loss function based on the Poisson distribution\footnote{Although we do not use it at this point, 
the Poisson loss also lends itself nicely for multiple target prediction which might be useful in e.g.\ subword modeling.}: we consider each $t_j$ 
in $t$ to be Poisson distributed with parameter $y_j$. The conditional probability of $P_{Poisson}(\mathbf{t}|\mathbf{f})$ then is:
\begin{equation}
	%\mathcal{L}_{Poisson}(t) = \prod_{i \in t}\frac{y_{i}^{t_i} e^{-y_i}}{t_i!}
	P_{Poisson}(\mathbf{t}|\mathbf{f}) = \prod_{j \in \mathbf{t}}\frac{y_{j}^{t_j} e^{-y_j}}{t_j!}
\end{equation}
and the corresponding Poisson loss function is:
\begin{align}
	%L_{Poisson}(y,t) = -\sum_{i \in t}[t_i\log(y_i) - y_i - \log(t_i!)]
	L_{Poisson}(\mathbf{y},\mathbf{t}) = -log(P_{Poisson}(\mathbf{t}|\mathbf{f})) &= -\sum_{j \in \mathbf{t}}[t_j\log(y_j) - y_j - log(t_j!)] \nonumber \\
										      &= \sum_{j \in \mathbf{t}}y_j -\sum_{j \in \mathbf{t}} t_j\log(y_j)
\end{align}
where we dropped the last term, since $t_j$ is binary-valued\footnote{In fact, even in the general case where $t_k$ can take any non-negative value, this term will disappear in the gradient, as it is independent of $\mathbf{M}$.}.
Although this choice is not obvious in the context of language modeling, it is well suited to gradient-based optimization and, as we will see, the experimental results are in fact excellent.
%\subsection{Adjustment Function Estimation}
\subsection{Model Estimation}

The adjustment function is learned by applying stochastic gradient descent on the 
loss function. That is, for each feature-target pair $(f_i,t_j)$ in each event 
%loss function. That is, for each event 
we need to update the parameters of the metafeatures by calculating the gradient with respect to 
the adjustment function.

For the multinomial loss, this gradient is:
%\begin{eqnarray}
\begin{align}
%	\frac{\partial log P(t_j|\mathbf{f})}{\partial \theta_k} &=\frac{1}{P(t_j|\mathbf{f})}\frac{\partial P(t_j|\mathbf{f})}{\partial \theta_k}
%	= \frac{\sum_{i \in \mathbf{f}} C_{ij} a(i,j)}{\sum_{i \in \mathbf{f}} \sum_{u \in \mathcal{V}} C_{iu} a(i,u)} \nonumber \\
%	&= \frac{\sum_{i \in \mathbf{f}} C_{ij} \frac{\partial a(i,j)}{\partial \theta_k}}{\sum_{i \in \mathbf{f}} C_{ij} a(i,j)} - 
%	\frac{\sum_{i \in \mathbf{f}} \sum_{u \in \mathcal{V}} C_{iu} \frac{\partial a(i,u)}{\partial \theta_k}}{\sum_{i \in \mathbf{f}} \sum_{u \in \mathcal{V}} C_{iu} a(i,u)}
	\frac{\partial(L_{multi}(\mathbf{Mf},\mathbf{t}))}{\partial(A(i,j))}
	&= \frac{\partial(log(\sum_{u=1}^{|\mathcal{V}|}(\mathbf{Mf})_u)-log(\mathbf{Mf})_j)}{\partial(M_{ij})} \frac{\partial(M_{ij})}{\partial(A_{ij})} \nonumber \\
	&= [\frac{\partial(log(\sum_{u=1}^{|\mathcal{V}|}(\mathbf{Mf})_u))}{\partial(M_{ij})} - \frac{\partial(log(\mathbf{Mf})_j)}{\partial(M_{ij})}] M_{ij} \nonumber \\
	&= [\frac{\partial(\sum_{u=1}^{|\mathcal{V}|}(\mathbf{Mf})_u)}{\sum_{u=1}^{|\mathcal{V}|}(\mathbf{Mf})_u \partial(M_{ij})} - \frac{\partial(\mathbf{Mf})_j}{(\mathbf{Mf})_j \partial(M_{ij})}] M_{ij} \nonumber \\
	&= (\frac{f_i}{\sum_{u=1}^{|\mathcal{V}|}(\mathbf{Mf})_u} - \frac{f_i \cdot t_j}{y_j}) M_{ij} \nonumber \\
	&= f_i M_{ij} (\frac{1}{\sum_{u=1}^{|\mathcal{V}|}y_u} - \frac{t_j}{y_j})
\end{align}
%\end{eqnarray}

The problem with this update rule is that we need to sum over the entire vocabulary $\mathcal{V}$ in %both the numerator and 
the denominator. For most features $f_i$, this is not a big deal as $C_{iu} = 0$, but some features occur with many if not 
all targets e.g. the empty feature for unigrams. Although we might be able to get away with this by re-using these sums and applying 
them to many/all events in a mini batch, we chose to work with the Poisson loss in our first implementation.

If we calculate the gradient of the Poisson loss, we get the following:

\begin{align}
\label{gradient}
%\frac{d(L_{Poisson}(Mf,t))}{d(A(f_i,t_j))} = f_i M_{ij} - \frac{f_it_jM_{ij}}{y_j}
\frac{\partial(L_{Poisson}(\mathbf{Mf},\mathbf{t}))}{\partial(A(i,j))} 
	&= \frac{\partial(\sum_{u=1}^{|\mathcal{V}|} \mathbf{(Mf)}_u - \sum_{u=1}^{|\mathcal{V}|} t_u \log(\mathbf{Mf})_u)}{\partial(M_{ij})} \frac{\partial(M_{ij})}{\partial(A(i,j))} \nonumber \\
	&= [\frac{\partial(\sum_{u=1}^{|\mathcal{V}|} \mathbf{(Mf)}_u)}{\partial(M_{ij})} - \frac{\partial(\sum_{u=1}^{|\mathcal{V}|} t_u\log(\mathbf{Mf})_u)}{\partial(M_{ij})} ] M_{ij} \nonumber \\
	&= [f_i - \frac{t_j}{(\mathbf{Mf})_j} \frac{\partial(\mathbf{Mf})_j}{\partial(M_{ij})}] M_{ij} \nonumber \\
	&= [f_i - \frac{t_j f_i}{(\mathbf{Mf})_j}] M_{ij} \nonumber \\
	&= f_i M_{ij} (1 - \frac{t_j}{y_j})
\end{align}

If we were to apply this gradient to each (positive and negative) training example, it would be computationally too expensive, 
because even though the second term is zero for all the negative 
training examples, the first term needs to be computed for all $|E| |Pos(\mathbf{f})| |\mathcal V|$ training examples. 

However, since the first term does not depend on $y_j$, we are able to distribute the updates for the negative examples over the 
positive ones by adding in gradients for a fraction of the events where $f_i = 1$, but $t_j = 0$. 
In particular, instead of adding the term $f_i M_{ij}$, we add $f_i t_j \frac{C_{i*}}{C_{ij}} M_{ij}$:
\begin{equation}
\frac{C_{i*}}{C_{ij}} M_{ij} \sum_{e=(f_i,t_j) \in E} f_i t_j = \frac{C_{i*}}{C_{ij}} M_{ij} C_{ij} = M_{ij} \sum_{e=(f_i,t_j) \in E} f_i
\end{equation}
which lets us update the gradient only on positive examples. We note that this update is only strictly correct for batch training, and not for online training since $M_{ij}$ changes after each update. Nonetheless, we found this to 
yield good results as well as seriously reducing the computational cost.
The online gradient applied to each training example then becomes: 
\begin{equation}
%\frac{d(L_{Poisson}(Mf,t))}{d(A(f_i,t_j))} = f_i t_j \frac{C_{i*}}{C_{ij}} M_{ij} - \frac{t_jf_iM_{ij}}{y_i}
\frac{\partial(L_{Poisson}(\mathbf{Mf},\mathbf{t}))}{\partial(A(i,j))} = f_i t_j M_{ij} (\frac{C_{i*}}{C_{ij}} - \frac{1}{y_j})
%\frac{d(L_{Poisson}(Mf,t))}{d(A(f_i,t_j))} = f_i t_j M_{ij} \frac{\frac{C_{i*}}{C_{ij}} y_i-1}{y_i}
\end{equation}
which is non-zero only for positive training examples, hence speeding up computation by a factor of $|\mathcal{V}|$.

These aggregated gradients however do not allow us to use additional data to train 
the adjustment function, since they tie the update computation to the relative frequencies $\frac{C_{i*}}{C_{ij}}$. Instead, we have to resort to leave-one-out training to prevent the model from 
overfitting the training data. We do this by excluding the event, generating the gradients, from the counts 
used to compute those gradients. So, for each positive example $(f_i,t_j)$ of each event $e = (\mathbf{f},\mathbf{t})$, we compute 
the gradient, excluding $f_i$ from $C_{i*}$ and $f_i t_j$ from $C_{ij}$. For the gradients of the negative examples 
on the other hand we only exclude $f_i$ from $C_{i*}$ and we leave $C_{ij}$ untouched, since here we did not observe $t_j$. 
In order to keep the aggregate computation of the gradients for the negative examples, we distribute them uniformly over all the positive examples 
with the same feature; each of the $C_{ij}$ positive examples will then compute the gradient of $\frac{C_{i*} - C_{ij}}{C_{ij}}$ 
negative examples.

To summarize, when we do leave-one-out training we apply the following gradient update rule on all positive training examples:
\begin{align}
\label{gradient_agg}
\frac{\partial(L_{Poisson}(\mathbf{Mf},\mathbf{t}))}{\partial(A(i,j))} &= 
%	f_i t_j \frac{C_{i*} - C_{ij}}{C_{ij}} \frac{C_{ij}}{C_{i*}-1}e^{\sum_k \theta(hash[\alpha_k(i,j,C_{i*}-1,C_{ij})])} \nonumber \\
%	&\; + f_i t_j \frac{C_{ij}-1}{C_{i*}-1} \frac{y_j-1}{y_j} e^{\sum_k \theta(hash[\alpha_k(i,j,C_{i*}-1,C_{ij}-1)])} 
	f_i t_j \frac{C_{i*} - C_{ij}}{C_{ij}} \frac{C_{ij}}{C_{i*}-1}e^{\sum_k \theta(hash[\alpha_k(i,j,C_{i*}-1,C_{ij})])} \nonumber \\
	&\; + f_i t_j \frac{C_{ij}-1}{C_{i*}-1} \frac{y'_j-1}{y'_j} e^{\sum_k \theta(hash[\alpha_k(i,j,C_{i*}-1,C_{ij}-1)])} 
%	&=f_i t_j \frac{C_{i*} - C_{ij}}{C_{i*}-1} e^{\sum_k \theta(hash[\alpha_k(i,j,C_{i*}-1,C_{ij})])} \nonumber \\
%	&\; + f_i t_j \frac{C_{ij}-1}{C_{i*}-1} \frac{y_j-1}{y_j} e^{\sum_k \theta(hash[\alpha_k(i,j,C_{i*}-1,C_{ij}-1)])} 
\end{align}
where $y'_j$ is the product of leaving one out for all the relevant features i.e.\ $y'_j = (\mathbf{M}'\mathbf{f})_j$ and 
$\mathbf{M}'_{ij} = e^{\sum_k \theta(hash[\alpha_k(i,j,C_{i*}-1,C_{ij}-1)])} \frac{C_{ij}-1}{C_{i*}-1}$.

\section{Experiments}
\label{sec:exp}

\subsection{Corpus: One Billion Benchmark}

Our experimental setup used the One Billion Word Benchmark corpus\footnote{http://www.statmt.org/lm-benchmark} made available by~\cite{Chelba:2013}.

For completeness, here is a short description of the corpus, containing only monolingual English data:
\begin{itemize}
\item Total number of training tokens is about 0.8 billion
\item The vocabulary provided consists of 793471 words including sentence boundary markers \verb+<S>+, \verb+<\S>+, and was constructed by discarding all words with count below 3
\item Words outside of the vocabulary were mapped to \verb+<UNK>+ token, also part of the vocabulary
\item Sentence order was randomized
\item The test data consisted of 159658 words (without counting the sentence beginning marker \verb+<S>+ which is never predicted by the language model)
\item The out-of-vocabulary (OoV) rate on the test set was 0.28\%.
\end{itemize}

\subsection{SNM for n-gram LMs}
When trained using solely n-gram features, SNM comes very close to the state-of-the-art Kneser-Ney~\cite{Kneser:1995} (KN) models. Table~\ref{tab:n-gram} 
shows that Katz~\cite{Katz:1987} performs considerably worse than both SNM and KN which only differ by about 5\%. When we interpolate these two models linearly,  
the added gain is only about 1\%, suggesting that they are approximately modeling the same things. The difference between KN and SNM 
becomes smaller when we increase the size of the context, going from 5\% for 5-grams to 3\% for 8-grams, which indicates that SNM is 
better suited to a large number of features.

\begin{table}
\centering
\begin{tabular}{|l|c|c|c|c|}
\hline
Model	&   5  &   6  &   7  &   8 \\
\hline
KN 	& 67.6 & 64.3 & 63.2 & 62.9\\
Katz	& 79.9 & 80.5 & 82.2 & 83.5\\
SNM	& 70.8 & 67.0 & 65.4 & 64.8\\
\hline
KN+SNM  & 66.5 & 63.0 & 61.7 & 61.4\\
\hline
\end{tabular}
\caption{Perplexity results for Kneser-Ney, Katz and SNM, as well as for the linear interpolation of Kneser-Ney and SNM. 
Optimal interpolation weights are always around $0.6-0.7$ (KN) and $0.3-0.4$ (SNM).}
\label{tab:n-gram}
\end{table}

\subsection{Sparse Non-negative Modeling for Skip $n$-grams}
\label{sec:snm_skip}
When we incorporate skip-gram features, we can either build a `pure' skip-gram SNM that contains no regular $n$-gram features, 
except for unigrams, and interpolate this model with KN, or we can build a single SNM that has both the regular $n$-gram features 
and the skip-gram features. We compared the two approaches by choosing skip-gram features that can be considered the skip-equivalent 
of 5-grams i.e.\ they contain at most 4 words. In particular, we used skip-gram features where the remote span is limited to at most 
3 words for skips of length between 1 and 3 ($r = [1..3]$, $s = [1..3]$, $r+a = [1..4]$) and where all skips longer than 4 are tied 
and limited by a remote span length of at most 2 words ($r = [1..2]$, $s = [4..*]$, $r+a = [1..4]$). We then built a model that uses 
both these features and regular 5-grams (SNM5-skip), as well as one that only uses the skip-gram features (SNM5-skip (no n-grams)).

As it turns out and as can be seen from Table~\ref{tab:skip-gram}, it is better to incorporate all the features into one single 
SNM model than to interpolate with a KN 5-gram model (KN5). Interpolating the all-in-one SNM5-skip with KN5 yields almost no additional gain.

\begin{table}
\centering
\begin{tabular}{|l|c|c|}
\hline
Model				& Num. Params 	& PPL  \\
\hline
SNM5-skip (no n-grams)		& 61 B		& 69.8 \\
SNM5-skip			& 62 B		& 54.2 \\
KN5+SNM5-skip (no n-grams)	&		& 56.5 \\
KN5+SNM5-skip			&		& 53.6 \\
\hline
\end{tabular}
\caption{Number of parameters (in billions) and perplexity results for SNM5-skip models with and without n-grams, as well as perplexity results for the interpolation with KN5.}
\label{tab:skip-gram}
\end{table}

The best SNM results so far (SNM10-skip) were achieved using 10-grams, together with untied skip features of at most 5 words with a skip of 
exactly 1 word ($s = 1$, $r+a = [1..5]$) as well as tied skip features of at most 4 words where only 1 word is remote, but up to 10 words can be 
skipped ($r = 1$, $s = [1..10]$, $r+a = [1..4]$).

This mixture of rich short-distance and shallow long-distance features enables the model to achieve state-of-the-art results, as 
can be seen in Table~\ref{tab:best}. When we 
compare the perplexity of this model with the state-of-the art RNN results in~\cite{Chelba:2013}, the difference is 
only 3\%. Moreover, although our model has more parameters than the RNN (33 vs 20 billion), training takes about a tenth of the time (24 hours vs 240 hours). Interestingly, when we interpolate the two models, we have an additional gain of 20\%, and as far as we know, the perplexity of 41.3 is 
already the best ever reported on this database, beating the previous best by 6\%~\cite{Chelba:2013}. 

Finally, when we optimize interpolation weights over all models in~\cite{Chelba:2013}, including SNM5-skip and SNM10-skip, the contribution of the other 
models as well as the perplexity reduction is negligible, as can be seen in Table~\ref{tab:best}, which also summarizes the perplexity results for each 
of the individual models.
%
%\begin{table}
%\centering
%\begin{tabular}{|l|c|}
%\hline
%		& PPL  \\
%\hline
%SNM10		& 52.9 \\
%RNN1024		& 51.3 \\
%SNM10+RNN1024	& 41.3 \\
%\hline
%\end{tabular}
%\caption{Comparison of SNM with state-of-the-art RNN and their linear interpolation. Optimal interpolation weights are 0.4 (SNM10) 
%and 0.6 (RNN1024).}
%\label{tab:rnn}
%\end{table}

\begin{table}
\centering
%\begin{tabular}{|l|c|c|}
\begin{tabular}{|l|c|c|c|c|c|}
\hline
%		& Weight & PPL  \\
Model			& Num. Params	& PPL	&\multicolumn{3}{|c|}{interpolation weights}  \\
\hline
KN5			& 1.76 B		& 67.6 	&	& 0.06	& 0.00	\\
HSME			& 6 B		& 101.3	&	& 0.00	& 0.00	\\
SBO			& 1.13 B	& 87.9 	&	& 0.20	& 0.04	\\
SNM5-skip		& 62 B		& 54.2 	&	& 	& 0.10	\\
SNM10-skip		& 33 B		& 52.9 	& 0.4	&	& 0.27	\\
RNN256			& 20 B		& 58.2 	&	& 0.00	& 0.00	\\
RNN512			& 20 B		& 54.6 	&	& 0.13	& 0.07	\\
RNN1024			& 20 B		& 51.3 	& 0.6	& 0.61	& 0.53	\\
\hline
SNM10-skip+RNN1024	&		& 	& 41.3	& 	&	\\
\hline
Previous best		&		& 	&	& 43.8	&	\\
\hline
ALL			&		& 	&	&	& 41.0 \\
\hline
\end{tabular}
\caption{Number of parameters (in billions) and perplexity results for each of the models in~\cite{Chelba:2013}, 
and SNM5-skip and SNM10-skip, as well as interpolation results and weights. }
\label{tab:best}
\end{table}

\subsection{Ablation Experiments}
\label{sec:ablations}
To find out how much, if anything at all, each metafeature contributes to the adjustment function, we ran a series 
of ablation experiments in which we ablated one metafeature at a time. When we experimented on SNM5, we found, 
unsurprisingly, that the most important metafeature is the feature-target count. At first glance, it does not seem 
to matter much whether the counts are stored in 1 or 2 buckets, but the second bucket really starts to pay off for 
models with a large number of singleton features e.g. SNM10-skip\footnote{Ideally we want to have the SNM10-skip ablation results
as well, but this takes up a lot of time, during which other development is hindered.}. 
This is not the case for the feature counts, where having a single bucket is always better, although in general the feature counts 
do not contribute much. In any case, feature counts are definitely the least important for the model. 
The remaining metafeatures all contribute more or less equally, all of which can be seen in Table~\ref{tab:ablations}.  

\begin{table}
\centering
\begin{tabular}{|l|c|}
\hline
Ablated feature	 		& PPL  \\
\hline
No ablation			& 70.8 \\
Feature				& 71.9 \\
Feature type			& 71.4 \\
Feature count			& 70.6 \\
Feature count: second bucket 	& 70.3 \\
Link count	 		& 73.2 \\
Link count: second bucket 	& 70.6 \\
\hline
\end{tabular}
\caption{Metafeature ablation experiments on SNM5}
\label{tab:ablations}
\end{table}

\section{Related Work}
\label{sec:related}

SNM estimation is closely related to all $n$-gram LM smoothing techniques that
rely on mixing relative frequencies at various orders. Unlike most of those, 
it combines the predictors at various orders without relying on a hierarchical 
nesting of the contexts, setting it closer to the family of maximum entropy
(ME)~\cite{Rosenfeld:1994}, or exponential models.

We are not the first ones to highlight the effectiveness of skip $n$-grams at 
capturing dependencies across longer contexts, similar to RNN LMs; previous such 
results were reported in \cite{Klakow:2013}. 

\cite{Chelba:2000} attempts to capture long range dependencies in language where
the skip $n$-grams are identified using a left-to-right syntactic parser. Approaches such as \cite{Bellegarda:2000} leverage latent semantic information, whereas \cite{Wang:2012} integrates both syntactic and topic-based modeling in a unified approach.

The speed-ups to ME, and RNN LM training provided by hierarchically predicting words 
at the output layer~\cite{Goodman:2001b}, and subsampling~\cite{Xu:2011} still require 
updates that are linear in the vocabulary size times the number of words in the training 
data, whereas the SNM updates in Eq.~(\ref{gradient_agg}) for the much smaller adjustment
function eliminate the dependency on the vocabulary size. Scaling up RNN LM training is described in~\cite{Chelba:2013} and \cite{Robinson:2015}.

The computational advantages of SNM over both Maximum Entropy and RNN LM estimation are 
probably its main strength, promising an approach that has 
the same flexibility in combining arbitrary features effectively and yet should scale 
to very large amounts of data as gracefully as $n$-gram LMs do.

\section{Conclusions and Future Work}
\label{sec:conclusions}
We have presented SNM, a new family of LM estimation techniques. A first empirical evaluation on the One Billion Word Benchmark~\cite{Chelba:2013} shows that SNM $n$-gram LMs perform almost as well as the well-established KN models. 

When using skip-gram features the models are able to match the stat-of-the-art RNN LMs; combining the two modeling techniques yields the best known result on the benchmark.

Future work items include model pruning, exploring richer features similar to~\cite{Goodman:2001a}, as well as richer metafeatures in the adjustment model, mixing SNM models trained on various data sources such that they perform best on a given development set, and estimation techniques that are more flexible in this respect.

\end{document}